\documentclass{ieeeaccess}
\usepackage{cite}
\usepackage{amsmath,amssymb,amsfonts}
\usepackage{algorithmic}
\usepackage{graphicx}
\usepackage{textcomp}
\usepackage{subcaption}
\usepackage{algorithm}
\usepackage{algorithmic} 
\usepackage{comment}
\usepackage{setspace}
\usepackage{makecell}
\usepackage{subcaption}
\usepackage{bm}
\makeatletter
\AtBeginDocument{\DeclareMathVersion{bold}
\SetSymbolFont{operators}{bold}{T1}{times}{b}{n}
\SetSymbolFont{NewLetters}{bold}{T1}{times}{b}{it}
\SetMathAlphabet{\mathrm}{bold}{T1}{times}{b}{n}
\SetMathAlphabet{\mathit}{bold}{T1}{times}{b}{it}
\SetMathAlphabet{\mathbf}{bold}{T1}{times}{b}{n}
\SetMathAlphabet{\mathtt}{bold}{OT1}{pcr}{b}{n}
\SetSymbolFont{symbols}{bold}{OMS}{cmsy}{b}{n}
\renewcommand\boldmath{\@nomath\boldmath\mathversion{bold}}}
\makeatother

\makeatletter
\providecommand{\@doi}{}
\providecommand{\doi}[1]{\gdef\@doi{#1}}
\makeatother

\def\BibTeX{{\rm B\kern-.05em{\sc i\kern-.025em b}\kern-.08em
    T\kern-.1667em\lower.7ex\hbox{E}\kern-.125emX}}

\begin{document}

\title{Discrete Prompt Tuning via Recursive Utilization of Black-box Multimodal Large Language Model for Personalized Visual Emotion Recognition}
\author{\uppercase{Ryo Takahashi}\authorrefmark{1}, \IEEEmembership{Student Member, IEEE}, 
\uppercase{Naoki Saito}\authorrefmark{2}, \IEEEmembership{Member, IEEE}, 
\uppercase{Keisuke Maeda}\authorrefmark{3}, \IEEEmembership{Member, IEEE}, 
\uppercase{Takahiro Ogawa}\authorrefmark{3}, \IEEEmembership{Senior Member, IEEE}, and 
\uppercase{Miki Haseyama}\authorrefmark{3}, \IEEEmembership{Senior Member, IEEE}
}
\address[1]{Graduate School of Information Science and Technology, Hokkaido University, Sapporo 060-0814, Japan}
\address[2]{Office of Institutional Research, Hokkaido University, Sapporo 060-0808, Japan}
\address[3]{Faculty of Information Science and Technology, Hokkaido University, Sapporo 060-0814, Japan}
\tfootnote{This work was partly supported by JSPS KAKENHI Grant Numbers JP24K02942, JP23K21676 and JP23K11211.}

\markboth
{R. Takahashi \headeretal: Discrete Prompt Tuning via Recursive Utilization}
{R. Takahashi \headeretal:Discrete Prompt Tuning via Recursive Utilization}

\corresp{Corresponding author: Miki Haseyama (e-mail: mhaseyama@lmd.ist.hokudai.ac.jp).}

\begin{abstract}
Visual Emotion Recognition~(VER)~is an important research topic due to its wide range of applications, including opinion mining and advertisement design. 
Extending this capability to recognize emotions at the individual level further broadens its potential applications.
Recently, Multimodal Large Language Models~(MLLMs)~have attracted increasing attention and demonstrated performance comparable to that
of conventional VER methods. 
However, MLLMs are trained on large and diverse datasets containing general opinions, which causes them to favor majority viewpoints and familiar patterns. 
This tendency limits their performance in a personalized VER, which is crucial for practical and real-world applications, and indicates a key area for improvement.
To address this limitation, the proposed method employs discrete prompt tuning inspired by the process of humans’ prompt engineering to adapt the VER task to each individual. 
Our method selects the best natural language representation from the generated prompts and uses it to update the prompt for the realization of accurate personalized VER.
\end{abstract}


\titlepgskip=-21pt

\maketitle

\section{Introduction}
\label{sec:introduction}
\PARstart{R}{ecognizing} human emotions is crucial for computers to respond appropriately to users' questions and requests.
In particular, research on human-induced emotion recognition by visual stimuli is typically addressed as Visual Emotion Recognition~(VER)~\cite{cnnrnn2017, xu2022mdan, xie2024emovit, zhu2025learning}.
VER is expected to have various applications, including advertisement~\cite{poels2006capture}, education~\cite{lopez2017mining}, mental health~\cite{guntuku2019twitter}, image retrieval~\cite{pang2015deep, guntuku2016likes}, and product recommendation~\cite{jaiswal2019intelligent}.
Although VER is a challenging research task because it deals with ambiguous information such as emotions, various methods have been proposed because of their vast applicability~\cite{moroto2023zero,kim2018building,li2019hierarchical}.

Recently, Multimodal Large Language Models~({MLLMs}) pretrained with numerous images and texts have demonstrated high performance on various tasks~\cite{wu2023can,zhang2024mm}. 
MLLMs are easy to use and can be applied to various tasks, which have been the subject of many studies. 
MLLMs can understand the relationship between different modalities, and various responses can be obtained by changing the prompt. 
In the VER, many MLLM-based VER methods have been proposed, achieving high performance~\cite{tzelepi2024disturbing,nadeem2024vision,lian2024gpt,bai2024m3d,zhang2024mathverse,lu2024gpt}. 

Emotional responses elicited by visual stimuli vary across individuals. 
For example, an image of a roller coaster may invoke feelings of ``excitement,'' in some viewers but “fear” in others. 
To apply VER technology to practical, real-world services, personalization of VER systems is essential. 
However, existing MLLM-based VER methods face challenges in generating recognition outcomes that reflect individual variability in emotional elicitation.
This is because MLLMs are trained on a large number of texts and images, which contain extensive general information and opinions; thus, they tend to learn majority opinions and common patterns.~\cite{mirchandani2023largelanguagemodelsgeneral,li2024formalityfavoredunravelinglearning,sheng-etal-2019-woman}. 

Fine-tuning is generally considered necessary to adapt an MLLM to a specific task. 
However, high-accuracy MLLMs tend to have numerous parameters, which poses the challenge of requiring substantial computational resources for fine-tuning.
To address this issue, prompt tuning is a potential approach for adapting MLLMs to specific tasks such as personalized VER. 
Unlike fine-tuning, prompt tuning does not modify the parameters of the MLLM. 
Instead, it adjusts the input prompts using techniques such as gradient-based optimization or soft prompt tuning so that the MLLM can perform effectively on specific tasks. 
Gradient-based prompt tuning~\cite{liu-etal-2022-p} is typically applied to white-box models, because these models allow access to model parameters and loss gradients.
However, recent MLLMs with superior reasoning capabilities, such as Gemini\footnote{\texttt{https://gemini.google.com/}} and GPT-4 omni (GPT-4o)\footnote{\texttt{https://openai.com/index/hello-gpt-4o/}}, are frequently deployed as black-box models, where access to parameters and loss gradients is restricted. 
This is primarily due to commercial, ethical, legal, security, and other concerns surrounding the release of model internals.
In prompt-based learning, an extensively used technique is ``soft prompts''~\cite{li2021prefix,gu2021ppt,lester2021power}. 
Soft prompts are learnable parameters represented as continuous vector embeddings that are prepended to the input of a pretrained language model. They are effective for improving model performance.
However, they are numerical and lack explicit semantics, making it difficult for humans to understand their meaning, particularly when interpreting decisions involving ambiguous emotions.
In addition, training with soft prompts requires access to gradient information. 
This poses a challenge in scenarios where users cannot access or modify the model internals, such as when interacting with models through restricted application programming interfaces or black-box systems.
To address these issues, researchers have proposed ``discrete prompt tuning'' approaches~\cite{liu2024language,diao2022black,pryzant-etal-2023-automatic}.
Discrete prompts comprise natural language tokens, which makes them more interpretable for humans and allows them to be used without relying on gradients. 
Therefore, there is a potential to adapt MLLMs to VER tasks by employing discrete prompt tuning, while maintaining interpretability and the adaptability of black-box models.

In this paper, we propose a personalized VER method that introduces discrete prompt tuning via recursive utilization of a black-box MLLM. 
We employ an MLLM capable of processing both images and text to perform the VER task, and use an LLM for prompt generation and refinement.
Specifically, the proposed method recognizes VER by inputting the target image and a discrete prompt into an MLLM. 
The discrete prompt is obtained via a prompt tuning approach that incorporates iterative generation and evaluation of discrete prompts inspired by the human prompt engineering process.
In the prompt generation phase, the LLM generates multiple prompts in natural language suitable for VER.
In the prompt evaluation phase, each generated prompt is input into the MLLM alongside training images for which the evoked emotions are known in advance. 
The recognition results for these images are then obtained, and the accuracy of these results is used as the performance score for each prompt. 
The proposed method then feeds the LLM with several high- and low-performing prompts and accuracy scores, which are obtained when these prompts are used to perform the VER task. 
Based on the information provided to the LLM, it generates new prompts that are expected to improve VER performance. 
The self-correction capability of the LLM enables the discovery of better prompts.
This process of prompt generation and evaluation is repeated to obtain prompts that are likely to enhance the accuracy of VER by the MLLM.
The proposed method leverages discrete prompt tuning to derive an optimal prompt for the VER for each user, which is expected to elicit knowledge related to VER in an MLLM according to the user's characteristics.
The discrete prompt tuning in the proposed method does not require changing the parameters of the MLLM and accessing its loss gradients; thus, it can be applied to both white-box and black-box MLLMs. 
This makes the proposed method less restrictive in its application than previously proposed MLLM-based VER methods.
Finally, the proposed method determines the final results through a majority vote. 
The MLLM processes the target images using several optimal prompts generated via discrete prompt tuning, and the most frequent label among these outputs is selected.

In summary, the main contributions of this paper are as follows.
\begin{itemize}
    \item We introduce a discrete prompt tuning approach for a black-box MLLM inspired by humans' prompt engineering to VER.  
    \item We use a discrete prompt tuning approach to improve personalized VER performance. 
    \item We achieve accurate VER by introducing a simple merging process of the VER results obtained from the MLLM by inputting several optimized prompts. 
\end{itemize}

This paper is organized as follows. 
Section~\ref{sec:Related works} describes related studies on VER and prompt-based learning. 
Section~\ref{sec:PM} explains the proposed method for personalized VER via discrete prompt tuning. 
In addition, experimental results are reported in Section~\ref{sec:experiment}. 
Finally, Section~\ref{sec:conclusions} concludes the paper.

\section{Related works}
\label{sec:Related works}
In this section, we explain related studies. 
The proposed method focuses on a VER task and a prompt-based learning approach for an MLLM. 
This section is organized as follows. 
Subsection~\ref{ssec:ver} introduces previous VER studies. 
Subsection~\ref{ssec:prompt} explains prompt-based learning approaches for MLLMs and their limitations. 

\subsection{VER}
\label{ssec:ver}
With the advancement of deep learning techniques, VER has significantly progressed.
In particular, Convolutional Neural Network~(CNN)-based approaches have emerged and significantly improved in recognition performance~\cite{peng2015mixed, you2016building}. 
For example, Chen et al.~\cite{chen2014deepsentibank} proposed DeepSentiBank, which classifies a visual sentiment via a deep CNN. 
This method automatically extracted adjective--noun pairs and was trained on approximately one million images. 
DeepSentiBank has several limitations, such as noisy and weak labels and domain adaptation challenges. 
You et al.~\cite{you2015robust} introduced PCNN, which combines progressive training and transfer learning to improve performance to deal with the limitations of DeepSentiBank.
In addition, She et al.~\cite{8825564} proposed WSCNet, a weakly supervised coupled network that integrates local regions evoking emotions with global image features. 
Focusing on the hierarchical structure of emotions, Xu et al.~\cite{xu2022mdan} presented MDAN, a multilevel dependent attention network built on CNNs, achieving fine-grained emotion classification via a hierarchy-aware learning strategy.

With the recent advancements in multimodal foundation models, various approaches leveraging these models have been proposed for VER, achieving recognition accuracies that exceed those of conventional methods based on CNNs and similar architectures~\cite{9964263,10388075,10889829,XU2024102366}. 
Among various foundation models, CLIP~\cite{DBLP:journals/corr/abs-2103-00020}, which learns from large-scale image–text pairs to align visual and linguistic representations in a common embedding space, has become a common backbone.
For instance, Deng et al.~\cite{deng2024learning} proposed a CLIP-based VER method that dynamically composes diverse prompts according to image content and emotion category, significantly outperforming fixed-prompt VER baselines.
Then, Deng et al. ~\cite{9964263} proposed a generation approach of knowledge-rich prompts by incorporating emotion words and entity-level information extracted from images. 
This approach enables more sophisticated affective understanding by leveraging linguistic structures beyond conventional label-based supervision.
In addition, MLLM-based VER methods such as GPT‑4o and Gemini have been proposed, and these methods have demonstrated high recognition performance even in zero‑shot settings for VER tasks~\cite{nadeem2024vision, lian2024gpt}.

However, MLLM-based VER methods cannot consider the individual differences in emotional elicitation for the realization of personalized VER. 
This is because MLLMs are trained on large and diverse datasets containing general opinions, causing them to favor majority viewpoints and familiar patterns. 
To overcome this limitation, a promising strategy is to adjust the parameters of MLLMs to enable specialization in personalized VER for each target user, i.e., a fine-tuning approach. 
However, fine-tuning these parameters requires a large number of training samples, and it is challenging to collect sufficient data specific to a target user's VER. 
Therefore, achieving effective fine-tuning for personalized VER remains difficult.
Our approach focuses on prompt tuning, which aims to improve VER performance by optimizing the instructions provided to the MLLM, rather than fine-tuning the model parameters. 
Unlike conventional fine-tuning, prompt tuning can achieve task-specific performance even with a limited amount of training data, making it a promising approach for realizing personalized VER using MLLMs. 
The following subsection explains related studies on prompt tuning of MLLMs. 

\subsection{Prompt tuning of MLLMs}
\label{ssec:prompt}
In recent years, many researchers have proposed prompt tuning methods as efficient alternatives to full fine-tuning, improving task performance while substantially reducing computational cost.
A notable gradient-based approach is P-Tuning, proposed by Liu et al.~\cite{liu-etal-2022-p}, which uses continuous trainable embeddings instead of discrete manually crafted prompts and optimizes them via backpropagation.
Similarly, Lester et al. ~\cite{lester2021power} introduced prompt tuning, which freezes the language model parameters and optimizes only a small set of task-specific soft prompts.
To further leverage gradient-based optimization for vision-language tasks, models such as InstructBLIP~\cite{dai2023instructblipgeneralpurposevisionlanguagemodels} and EmoVIT~\cite{xie2024emovit} have been proposed to align prompts with supervised instruction datasets.

However, the emergence of powerful black-box models, which do not reveal internal parameters or gradients, has made gradient-based tuning approaches impractical in many real-world scenarios.
In response, researchers have developed gradient-free prompt optimization techniques that function effectively in black-box settings.
For example, Pryzant et al.~\cite{pryzant-etal-2023-automatic} proposed ProTeGi, a method in which the language model generates natural language feedback to identify prompt weaknesses, thereby guiding its iterative refinement without relying on gradient signals.
Similarly,  Cheng et al.~\cite{cheng2024blackboxpromptoptimizationaligning} introduced black-box prompt optimization, which trains a sequence-to-sequence model to rewrite prompts by learning from examples of both effective and ineffective responses.
More recently, Liu et al.~\cite{liu2024language} proposed a dialogue-based framework that employs language models as interactive prompt engineers. 

We introduce a prompt tuning approach applicable to black-box models for VER.
The proposed method enables personalized VER by obtaining optimal prompts tailored to each individual.

\begin{figure}[t]
    \centering
    \includegraphics[width=90mm]{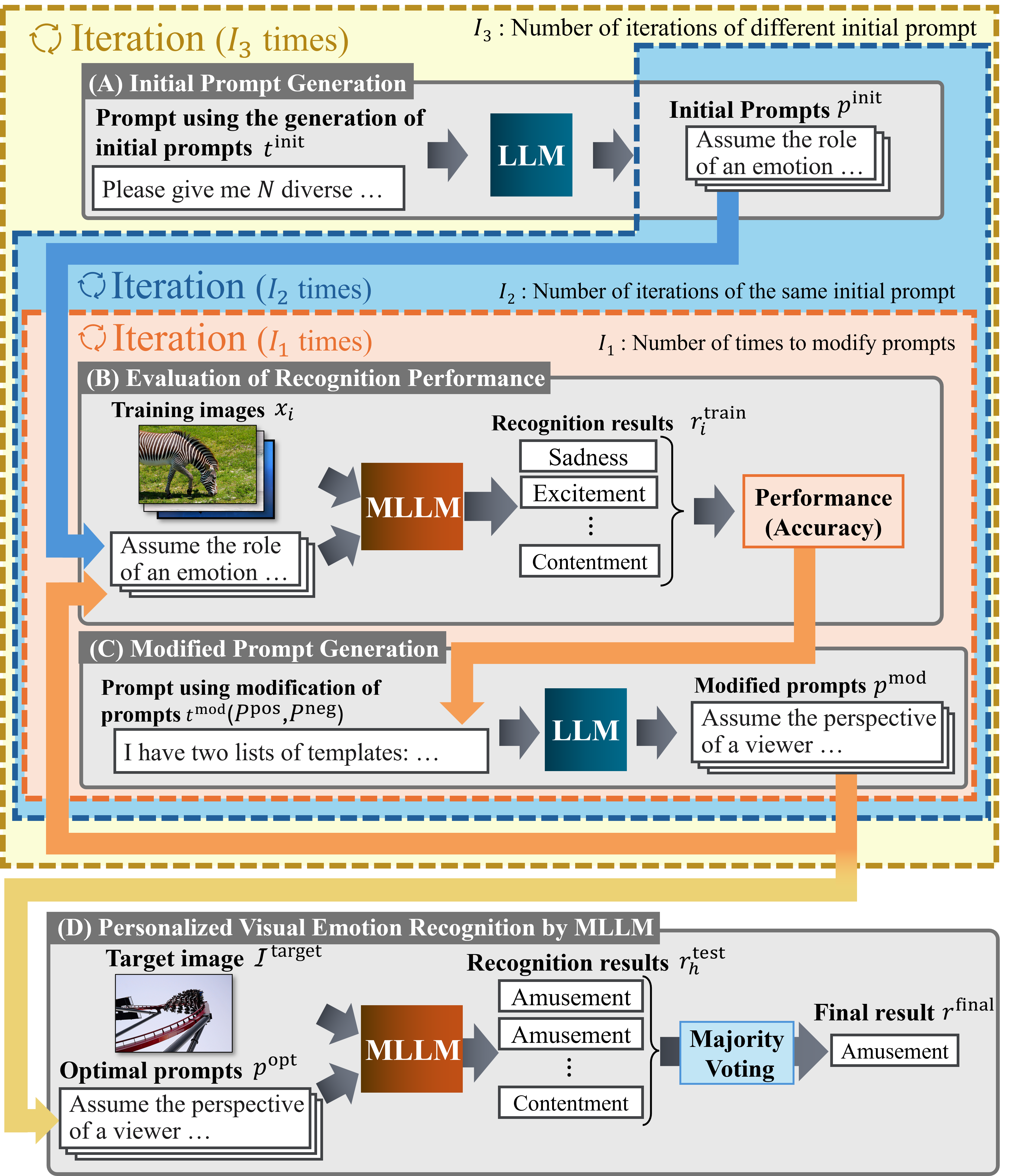}
    \caption{Overview of the proposed method. The proposed method comprises four processes: initial prompt generation, evaluation of recognition performance, modified prompt generation, and recognition by an MLLM. The proposed method iterates these processes and obtains the final recognition result. }
    \label{fig:overview}
\end{figure}

\begin{algorithm}[t]
\caption{VER via discrete prompt tuning of MLLM}
\textbf{Input}: Training samples $\{x_l, y_l\}_{l=1}^L$, number of times to modify prompts $I_1$, number of iterations of the same initial prompts $I_2$, number of iterations of different initial prompt $I_3$, number of initial prompts $N$, number of prompts to generate modified prompts $k$, target image $\mathcal{I}^{\text{target}}$, and number of recognition results to derive final result $H$.\\
\textbf{Output}: VER result $r^\text{final}$. 
\begin{algorithmic}[1]
    \STATE $P^{\text{all}} \gets \varnothing$
    \FOR{$1$ to $I_3$} 
        \STATE $P^{\text{init}} \gets \text{LLM}(t^\text{init})$
        \FOR{$1$ to $I_2$} 
            \STATE $P^{\text{rank}} \gets P^{\text{init}}$
            \FOR{$1$ to $I_1$} 
                \STATE $P^{\text{pos}} \gets$ prompts with $\text{top-}k$ $\text{ACC}(p)_{p \in P^{\text{rank}}}$
                \STATE $P^{\text{neg}} \gets$ prompts with $\text{worst-}k$ $\text{ACC}(p)_{p \in P^{\text{rank}}}$
                \STATE Get new prompts $P^{\text{mod}} \gets \text{LLM}(t^{\text{mod}}(P^{\text{pos}}, P^{\text{neg}}))$
                \STATE $P^{\text{rank}} \gets P^{\text{rank}} \cup P^{\text{mod}}$
            \ENDFOR
            \STATE $P^{\text{all}} \gets P^{\text{all}} \cup P^{\text{rank}}$
        \ENDFOR
    \ENDFOR
    \STATE $P^{\text{used}} \gets \varnothing$
    \FOR{$1$ to $H$}
        \STATE $P^{\text{all}} \gets P^{\text{all}} \setminus P^{\text{used}}$
        \STATE $p^{\text{opt}}_h \gets \arg \max_{p^{\text{all}}_m \in P^{\text{all}}} \text{ACC}(p^{\text{all}}_m)$
        \STATE $r^{\text{test}}_h \gets \text{MLLM}(\mathcal{I}^{\text{target}}, p^{\text{opt}}_h)$
        \STATE $P^{\text{used}} \gets P^{\text{used}} \cup p^{\text{opt}}_h$
    \ENDFOR
    \STATE $r^{\text{final}} \gets$ emotion label by majority voting from $\{r^{\text{test}}_1, r^{\text{test}}_2, \cdots, r^{\text{test}}_H\}$.
\end{algorithmic}
\end{algorithm}

\section{Personalized VER via Discrete Prompt Tuning}
\label{sec:PM}
In this section, we describe the proposed method.
The proposed method inputs images and user-specific discrete prompts optimized for emotion recognition into an MLLM.
Our method performs discrete prompt tuning based on the literature~\cite{liu2024language} inspired by the typical workflow of humans' prompt engineering process.
We recognize the personalized emotion from the target image using the MLLM, inputting the optimized prompt.
Figure~\ref{fig:overview} shows an overview of the proposed method.
The precise processes of the proposed method are shown in Algorithm~1.
The prompt tuning approach in the proposed method comprises the following iterations: (A) creating initial prompts for VER; (B) evaluating the recognition performance of prompts; (C) creating new modified prompts based on the prompt performance evaluation results obtained by (B), and then, repeatedly performing (B) and (C); (D) selecting the prompt with the highest recognition performance as the final discrete prompt.
The proposed method can generate discrete prompts that are suitable for the VER of each user using the above iterative workflow.
In addition, using multiple suitable prompts, the proposed method performs VER on the target image several times and obtains multiple recognition results.
Then, through majority voting from these results, one emotion label is selected as the final recognition result. 
Consequently, stable and highly accurate VER becomes feasible using the proposed method. 

This section is organized as follows.
Subsection~\ref{ssec: initial} explains the generation approach of initial prompts. 
Subsection~\ref{ssec: evaluation} describes the recognition performance evaluation of the generated prompts. 
Subsection~\ref{ssec: modified} presents the modification scheme of the generated prompts for personalized VER.
Subsection~\ref{ssec: classification} explains the recognition approach using the final discrete prompts.

\subsection{Initial Prompt Generation}
\label{ssec: initial}
In this subsection, we explain the generation approach of initial prompts. 
The proposed method constructs initial prompts to obtain suitable discrete prompts for personalized VER using an MLLM. 
Specifically, an LLM is used to generate the initial prompt set $P^{\text{init}} = \{ p^{\text{init}}_1, p^{\text{init}}_2, \cdots, p^{\text{init}}_N \}$ ($N$ being the number of initial prompts) as follows:
\begin{equation}
P^\text{init}=\text{LLM}(t^\text{init}),
\end{equation}
where $\text{LLM}(\cdot)$ denotes the LLM used for generating initial prompts and $t^{\text{init}}$ denotes the natural language sentence given to the LLM as a prompt to generate the initial prompts for VER. 
The proposed method uses the sentence $t^{\text{init}}$ shown in TABLE~\ref{tablePrompt}. 
To explore diverse instructional strategies, the LLM is guided to generate various prompts. 
Based on these initial prompts, the proposed method generates suitable prompts for personalized VER through the following sequential steps. 

\subsection{Evaluation of Recognition Performance}
\label{ssec: evaluation}
In this subsection, we describe the VER performance evaluation of the generated prompts. 
The proposed method evaluates the recognition performance of the generated prompts by measuring the recognition accuracy as a basis for generating prompts.
Let $S_L = \{x_l, y_l\} \quad (l = 1, 2, \dots, L; L$ being the number of training images) be the $l$-th training image for personalized VER, where $x_
l$ denotes the $l$-th training image. 
In addition, $y_l$ denotes the emotion label evoked when the target user sees the image $x_l$. 
The proposed method calculates the recognition accuracy $\text{ ACC}(p)$ defined in the following equation for VER using an MLLM with a prompt $p$. 
\begin{equation}
\text{ACC} ({p}) = \frac{1}{L} \sum_{l=1}^{L} {C}_l({p}),
\end{equation}
\begin{equation}
{C}_l ({p})=
\begin{cases}
1 & \text{if } r^{\text{train}}_l  ({p}) = y_l, \\
0 & \text{otherwise},
\end{cases}
\end{equation}
where $r^{\text{train}}_l ({p})$ denotes the recognition result of the $l$-th training image obtained using the MLLM with the prompt $p$.
Based on the recognition accuracy, the proposed method modifies the prompts to achieve high-performance VER, as discussed in the next subsection. 

\begin{table}[t]
    \centering
    \caption{Sentence $t^{\text{init}}$ using the generation of initial prompts in the proposed method.}
    \label{tablePrompt}
    \begin{tabular}{p{8cm}}
        \hline
Please provide me with \colorbox{yellow}{$N$} diverse prompts that are suitable for input into the MLLM. The prompts should be diverse, such as detailed and straightforward, and should give the LLM a role. The prompts should make the LLM classify the emotions that people evoke when they see the image. The emotion label should prompt the LLM to choose one of the following emotions: \colorbox{yellow}{\{\text{Emotion labels}\}}.
\\Here are my requirements:
\\\ - Please only reply with the template.
\\\ - Each template should start with `-' in a separate line.
\\\ - Ensure that the output is in a clear and consistent format.
\\
\hline
\end{tabular}
\end{table}

\begin{table}[t]
    \centering
    \caption{Prompt sentence $t^{\text{mod}}(P^{\text{pos}}, P^{\text{neg}})$ using the modification of prompts in the proposed method.}
    \label{tablePrompt2}
    \begin{tabular}{p{8cm}}
\hline
I have two lists of templates: one with good templates and the other with bad templates. Based on the characteristics that make a template good or bad, please provide $T$ better templates.
Here is the list of good templates with their accuracies:
\\Top-$k$: \colorbox{yellow}{[prompts with top-$k$ accuracy]}
\\Here is the list of bad templates with their accuracies:
\\Worst-$k$: \colorbox{yellow}{[prompts with worst-$k$ accuracy]}
\\Here are my requirements:
\\\ - Please only reply with the template.
\\\ - Each template should start with `-' in a separate line.
\\\ - Ensure that the output is in a clear and consistent format.
\\
\hline
\end{tabular}
\end{table}

\subsection{Modified Prompt Generation}
\label{ssec: modified}
This subsection describes the procedure for generating modified prompts based on the recognition accuracy obtained in the previous subsection. 
The proposed method generates modified discrete prompts that achieve high-performance personalized VER using an MLLM. 
First, the proposed method uses top-$k$ and bottom-$k$ prompts to improve recognition accuracy. 
Let $P^\text{pos}$ and $P^{\text{neg}}$ be the sets of appropriate and inappropriate prompts, separately selected from the previously generated set of prompts $P^\text{rank}$. 
The proposed method generates a new set of modified prompts $P^\text{mod}$ by inputting the prompt $t^{\text{mod}}(P^{\text{pos}}, P^{\text{neg}})$ to an LLM as follows:
\begin{equation}
P^\text{mod} = \text{LLM}(t^\text{mod}(P^\text{pos},P^\text{neg})),
\end{equation}
where $t^\text{mod}(P^\text{pos},P^\text{neg})$ denotes the prompt with $P^\text{pos}$ and $P^\text{neg}$. 
Specifically, the proposed method uses the natural language prompt $t^\text{mod}(P^\text{pos}, P^\text{neg})$ in TABLE~\ref{tablePrompt2}. 
Then, the proposed method obtains $T$ modified prompts, which are added to the set $P^\text{rank}$. 
Thus, the proposed method constructs modified prompts that facilitate more personalized VER using an MLLM.
The modification information includes both appropriate and inappropriate prompts. 
The LLM refines prompt representations by increasing and
decreasing their similarities to appropriate and inappropriate examples, respectively.
These implicit gradients indicate the direction toward the optimal prompt for the user, thereby enabling the efficient generation of prompts that suit each user.

\subsection{Personalized VER by MLLM}
\label{ssec: classification}
In this subsection, we explain the recognition approach using an MLLM inputting the optimal prompts. 
The proposed method iteratively generates prompts and performs VER using them to identify optimal prompts.
Specifically, the proposed method performs $I_1$ iterations of evaluating and modifying discrete prompts, and ultimately obtains the prompts with the highest accuracy using the training images.
In addition, after repeating the acquisition of prompts via evaluation and modification processes, these processes are repeated $I_2$ times using the same initial prompt by assigning $P^\text{init}$ to $P^\text{rank}$.
Then, these processes are repeated $I_3$ times by generating new initial prompts $P^\text{init}$.
Note that the generated prompts are denoted as $P^{\text{all}} = \{ p^\text{all}_1, p^\text{all}_2, \cdots, p^\text{all}_M \}$ ($M$ denoting the total number of prompts generated in all previous iterations).
The proposed method obtains the optimal prompt $ p^\text{opt}_h (h = 1, 2, \dots, H; H$ denoting the number of used prompts for the target) for the final VER task as follows:
\begin{equation}
p^\text{opt}_h = \arg \max_{p^{\text{all}}_m} \text{ACC}(p^{\text{all}}_m),
\end{equation}
where the prompt $p^\text{opt}_h$ is used to obtain the recognition result. 

In the proposed method, the recognition result $ r^{\text{test}}_h$~($h\in\{1, 2, \dots , H\}$)~for the target image $\mathcal{I}^\text{target}$ using an MLLM is as follows:
\begin{equation}
r^{\text{test}}_h = \text{MLLM}(\mathcal{I}^\text{target}, p^\text{opt}_h).
\end{equation}
The proposed method selects the optimal prompt $p^{\text{opt}}_{h+1}$ once again among the prompts $P^{\text{all}}$ excluding the prompt used $P^{\text{used}}$, and performs VER.
Multiple recognition results $\{ r^{\text{test}}_1, r^{\text{test}}_2, \cdots, r^{\text{test}}_H \}$ are obtained by repeating this process.
The emotion label obtained through the majority voting of these recognition results becomes the final recognition result. 
Consequently, the proposed method enables accurate and stable personalized VER.

\section{Experimental Results}
\label{sec:experiment}
\subsection{Experimental settings}
In this experiment, we verify the effectiveness of the proposed method for personalized VER.
This experiment evaluated the performance of the proposed method by classifying images based on the emotion labels assigned to each user.  
To evaluate personalized VER performance, we used the Affective Explanations (Affection) dataset~\cite{achlioptas2023affection}, which comprises images and emotion labels elicited from each viewer.
In the Affection dataset, each image was assigned eight types of emotion labels based on Mikel’s wheel from the psychological model~\cite{mikels2005emotional}: amusement, awe, contentment, excitement, anger, disgust, fear, and sadness. 
This experiment used images labeled with emotion labels from 15 users in the Affection dataset. 
These users were selected because they provided multiple images for each emotion label; individuals with either a small number of responses or an extremely large number of responses were excluded. 
Note that the average number of images used per user was 864, and the total number of used images was 12,964.
In practical use cases, it is challenging to obtain numerous image--emotion label sets from a specific user for VER because of the considerable effort required~\cite{korovina2019reliability, huang2024survey}. 
Therefore, this experiment assumed that emotion recognition would be performed using a few image--emotion label pairs as training data.
Specifically, we used 30\% of each user's image-emotion label pairs as training data and the remaining 70\% as test data. 

In this experiment, we employed the Large Language and Vision Assistant v1.6-Mistral-7B (LLaVA-v1.6-Mistral-7B)~\cite{liu2024visual} as the MLLM to recognize emotions and GPT-4o as the LLM to facilitate discrete prompt tuning.
In each iteration of the proposed method, this experiment used six initial prompts ($N=6$) and generated five modified prompts ($T=5$).
In addition, the number of prompts used for each of the top and worst inputs for $k$ was three.
Furthermore, $I_1$ = 20, $I_2$ = 2, $I_3$ = 3, and $H$ = 5.

To evaluate the effectiveness of the proposed method, its recognition performance was compared with that of three baseline methods~(CMs 1--3) and four ablation study cases~(CMs 4--7). 
CM1 is a CNN-based emotion recognition method~\cite{8825564}, and CM2 employs Transformer-based recognition~\cite{liu2021swintransformerhierarchicalvision}. 
A recognition method by an MLLM inputting a prompt proposed in the previous study~\cite{lian2024gpt}~is used as CM3, and CM4 is a recognition method that uses an initial prompt. 
CM5 is a recognition method using a prompt modified in one iteration, $i.e.$, $I_2 = I_3 = 1$, and CM6 is a recognition method using discrete prompts tuned to another user who is not the target user.
Additionally, CM7 is a recognition method that uses a single modified prompt without majority voting.

Three types of evaluation indices were used to evaluate VER performance: recognition accuracy, Emotion Confusion Confidence (ECC), and Emotional Misclassification Confidence (EMC). 
ECC and EMC are metrics proposed in~\cite{zhao2024to} for evaluating the performance of an emotion recognition method, utilizing the concept of emotional distance to mitigate misrecognition based on Mikel's wheel from psychological theory. 
The three evaluation indices are defined as follows:
\begin{equation}
\begin{split}
\text{Accuracy}&=\frac{N_\text{correct}}{N_\text{test}}\\&=\frac{\sum_{\alpha\in\text{labels}}\sum_{\beta\in\text{labels}}S_{\alpha \beta}\times M_{\alpha \beta}}{N_\text{test}},
\end{split}
\end{equation}
\begin{equation}
M_{\alpha \beta}=
\begin{cases}
1 & \text{if } \alpha = \beta, \\
0 & \text{otherwise},
\end{cases}
\end{equation}
\begin{equation}
\begin{split}
\text{ECC}&=\frac{\sum_{\alpha\in\text{labels}}\sum_{\beta\in\text{labels}}S_{\alpha \beta}\times\frac{1}{W_{\alpha \beta}}}{N_\text{test}}\\
&=\text{Accuracy}+\frac{\sum_{\alpha\in\text{labels}}\sum_{\beta\in\text{labels},\alpha\neq\beta}S_{\alpha \beta}\times\frac{1}{W_{\alpha \beta}}}{N_\text{test}},
\end{split}
\end{equation}
\begin{equation}
\text{EMC}=\frac{\sum_{\alpha\in\text{labels}}\sum_{\beta\in\text{labels},\alpha\neq\beta}S_{\alpha \beta}\times\frac{1}{W_{\alpha \beta}-1}}{N_\text{test}-N_\text{correct}},
\end{equation}
where $N_\text{test}$ and $N_\text{correct}$ denote the number of test images and correctly recognized images, respectively, $\text{labels}$ denote a set of the target emotion labels and $S_{\alpha\beta}$ represents the number of images recognized from the correct emotion label $\alpha$ to the label $\beta$. 
In addition, $W_{\alpha\beta}$ denotes the emotional distance, which is defined as follows:
\begin{equation}
W_{\alpha\beta}=
\begin{cases}
1+\text{dist}(\alpha,\beta) & \text{if } \alpha,\beta\in {C}_e({p}),\\
C+\text{dist}(\alpha,\beta) & \text{otherwise},
\end{cases}
\end{equation}
where $\text{dist}(\alpha,\beta)$ denotes the number of steps on Mikel's wheel, and $C$ represents a constant that scales the importance of polarity, and it is set to four to ensure that thedistance between emotions of the same polarity is always smaller than that of emotions with different polarities following the method established in~\cite{zhao2024to}. 
${C}_e({p})$ denotes a set of target emotion labels with the same polarity.
Note that VER task using an MLLM occasionally outputs non-targeted labels, such as ``happy'' and ``sad,'' which are considered incorrect results in this experiment. 

\begin{table*}[t]
    \centering
    \caption{Accuracy, ECC, and EMC of VER results obtained using proposed and comparative methods.}
    \label{table1}
    \begin{tabular}{lccc}  
        \hline
        & Accuracy & ECC & EMC\\ \hline 
        CM1 & 36.7\% $\pm$ 10.3\%  & 56.7\% $\pm$ 8.09\%& 54.0\% $\pm$ 8.20\%\\
        CM2 & 30.7\% $\pm$ 9.43\%  & 53.0\% $\pm$ 7.59\%  & 55.1\%$\pm$ 8.08\%  \\
        CM3 & 39.1\% $\pm$ 9.70\%  & 59.8\% $\pm$ 6.81\%& \textbf{59.9\% $\pm$ 7.90\%}\\
        CM4 & 40.6\% $\pm$ 8.27\%  & 60.1\% $\pm$ 5.93\%& 56.8\%$\pm$ 7.07\%\\
        CM5 & 41.6\% $\pm$ 9.18\%  & 60.9\% $\pm$ 6.34\%& 57.2\%$\pm$ 6.90\%\\
        CM6 & 40.4\% $\pm$ 8.64\%  & 59.9\%$\pm$ 6.12\% & 56.6\%$\pm$ 6.45\%\\
        CM7 & 43.1\% $\pm$ 8.63\%  & 61.8\% $\pm$ 6.05\%& 56.7\%$\pm$ 6.95\%\\
        Our method & \textbf{44.9\%} $\pm$ \textbf{9.62\%}  & \textbf{63.4\%} $\pm$ \textbf{6.61\%}& 58.8\%$\pm$ 7.92\%\\
        \hline
    \end{tabular}
\end{table*}

\subsection{Quantitative experimental results}
TABLE~\ref{table1} shows the mean and standard deviation of the accuracy, ECC, and EMC values calculated from the VER results performed for each user using the proposed and comparative methods.
These results demonstrate that the proposed method achieves the highest VER performance compared with the comparative methods.
The proposed method is also effective in scenarios in which the MLLM is a black box model and does not allow for soft prompt injection.
Compared with CM1 and CM2, both the baseline methods using MLLM and the proposed method outperform the CNN- and Transformer-based methods in terms of VER performance.
These results confirm that the proposed method can realize a high-performance VER compared with methods based on conventional deep learning architectures. 
The linguistic capabilities of MLLMs are assumed to contribute to the recognition of not only objects in images but also the relationships between objects and people, thereby enhancing VER.

\begin{figure}[t] 
    \centering
    \includegraphics[width=90mm]{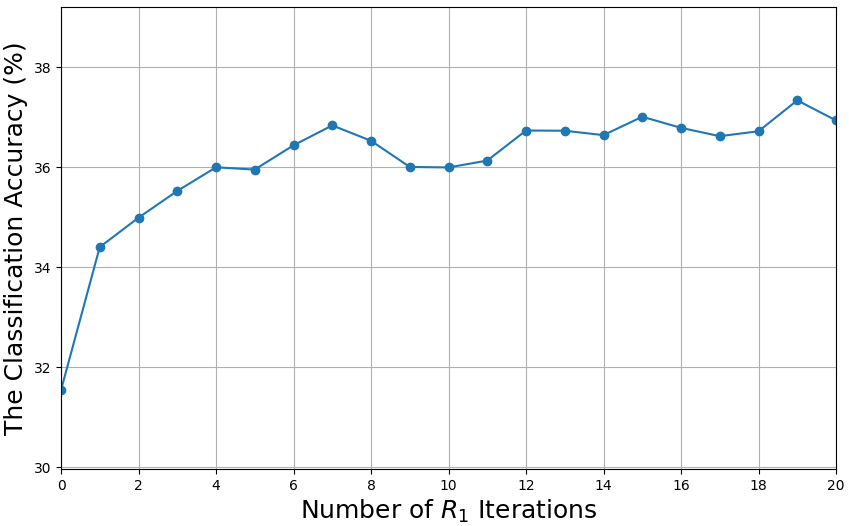} 
    \caption{Transition of recognition accuracy of training images by modifying prompts.}
    \label{fig:graph1}
\end{figure}

Compared with CMs 3--7, the proposed method, which applied prompt tuning to the MLLM, achieved the highest VER performance in terms of accuracy and ECC. 
These results demonstrate the importance of tailoring prompts to the MLLM.
In contrast, CM3 achieved the highest EMC score compared with the proposed method.
The EMC metric evaluates the plausibility of the predicted labels for images that were not correctly classified.
Some of these images may have features that make it easy to infer emotional labels.
This may have contributed to the relatively high EMC score of CM3.
Compared with CM4, the discrete prompt tuning approach of the
proposed method improves the performance by generating suitable prompts for personalized VER. 
Figure~\ref{fig:graph1} shows the transition of the recognition accuracy of the training samples by modifying the prompts. 
This transition confirms that the performance improvement in the
proposed method is not simply due to the selection of high-accuracy prompts from a set of randomly generated candidates but rather is attributable to the introduction of discrete prompt tuning.
As the number of iterations increases, the generated prompts evolve to incorporate features that are more suitable for each user, thereby improving accuracy.
Because the proposed method outperforms CM5, we can conclude that the iterative process of repeating the initial prompts achieves high-performance recognition. 
This is because the proposed method generates various prompts that are not fixed to a single pattern. 
Furthermore, compared with CM6, the proposed method generates suitable prompts for each user's emotion recognition. 
Therefore, we can confirm that the prompts generated by the proposed method are personalized, rather than generic for VER users in general. 
Finally, a comparison of the emotion recognition results of the proposed method and CM7 verifies that the majority voting approach effectively obtains the final recognition results. 
Using the majority voting scheme, the proposed method obtains the final recognition results by considering various prompts. 
Consequently, the influence of prediction errors and outliers by multiple prompts is averaged out, enabling the proposed method to derive stable and robust recognition results. 
The proposed approach ultimately results in improved recognition accuracy.

\begin{table*}[!t]
    \centering
    \caption{Optimal prompts for each iteration by proposed method.}
    \label{table2}
    \begin{tabular}{l l c p{90mm}}
        \\ \hline
        &Iteration conditions & Accuracy & Prompt sentences $p^\text{opt}_1$\\
         \hline
        Prompt (\romannumeral 1) & \makecell[tl]{$I_1$ = 1, $I_2$ = 1, $I_3$ = 1\\(Initial prompt)} & 39.4\% & Given the image, identify the primary emotion it evokes: \{amusement, awe, contentment, excitement, anger, disgust, fear, sadness\}. \\\hline
        Prompt (\romannumeral 2) & $I_1$ = 10, $I_2$ = 1, $I_3$ = 1 & 42.8\% & As a professional in emotion analysis, assess this image and choose the most prominent emotion it conveys: \{amusement, awe, contentment, excitement, anger, disgust, fear, sadness\}. Provide a short justification for your choice. \\\hline
        Prompt (\romannumeral 3) & $I_1$ = 20, $I_2$ = 1, $I_3$ = 1 & 43.9\% & As a professional in emotional interpretation, scrutinize this image and determine the most evident emotion it projects: \{amusement, awe, contentment, excitement, anger, disgust, fear, sadness\}. Offer a concise justification for your decision. \\\hline
        Prompt (\romannumeral 4) & \makecell[tl]{$I_1$ = 20, $I_2$ = 2, $I_3$ = 3\\(Final prompt)} & \textbf{45.4\%} & \textbf{As a choreographer expressing feelings through movement, recognize the emotion most evident in this image. Select from \{amusement, awe, contentment, excitement, anger, disgust, fear, sadness\}. }
        \\ \hline
    \end{tabular}
\end{table*}

\begin{figure*}[t] 
    \centering
    \includegraphics[width=150mm]{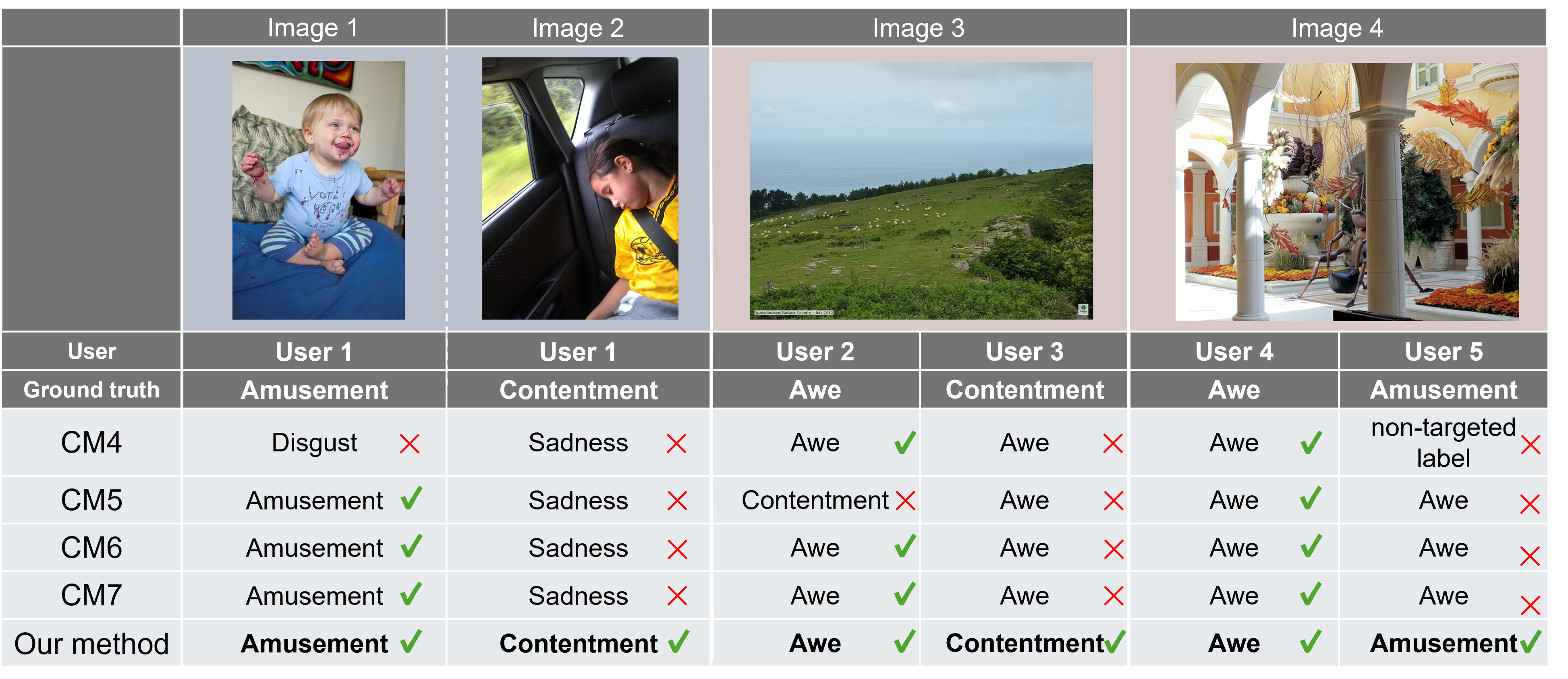} 
    \caption{Examples of target images and their VER results by proposed and comparative methods.}
    \label{fig:example1}
\end{figure*}


\begin{figure*}[htbp]
  \centering
  \begin{subfigure}[b]{0.49\linewidth}
    \centering
    \includegraphics[width=\linewidth]{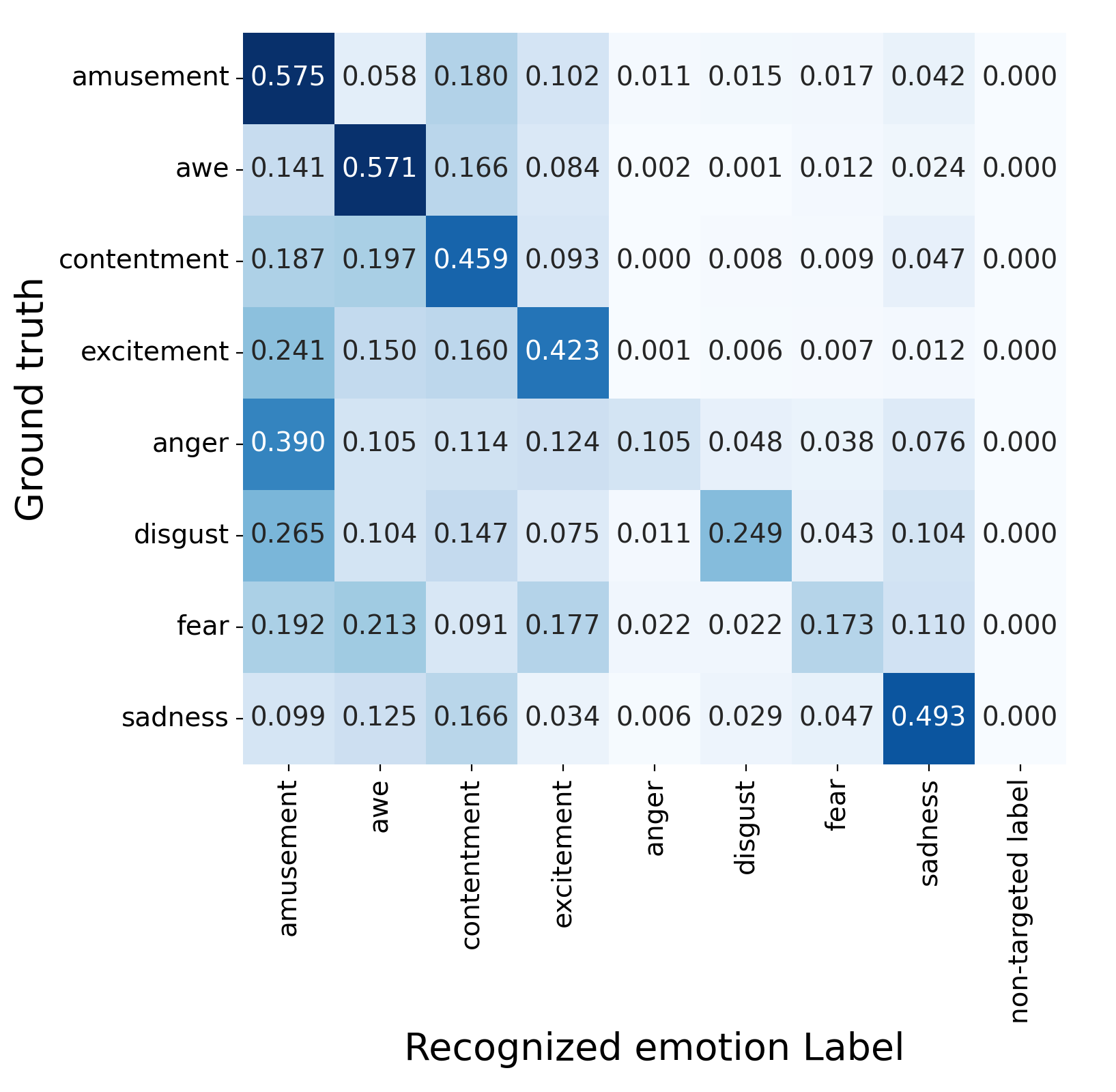}
    \vspace{-3mm}
    \caption{(i) Our method}
    \label{fig:matrix_a}
  \end{subfigure}
  \hfill
  \begin{subfigure}[b]{0.49\linewidth}
    \centering
    \includegraphics[width=\linewidth]{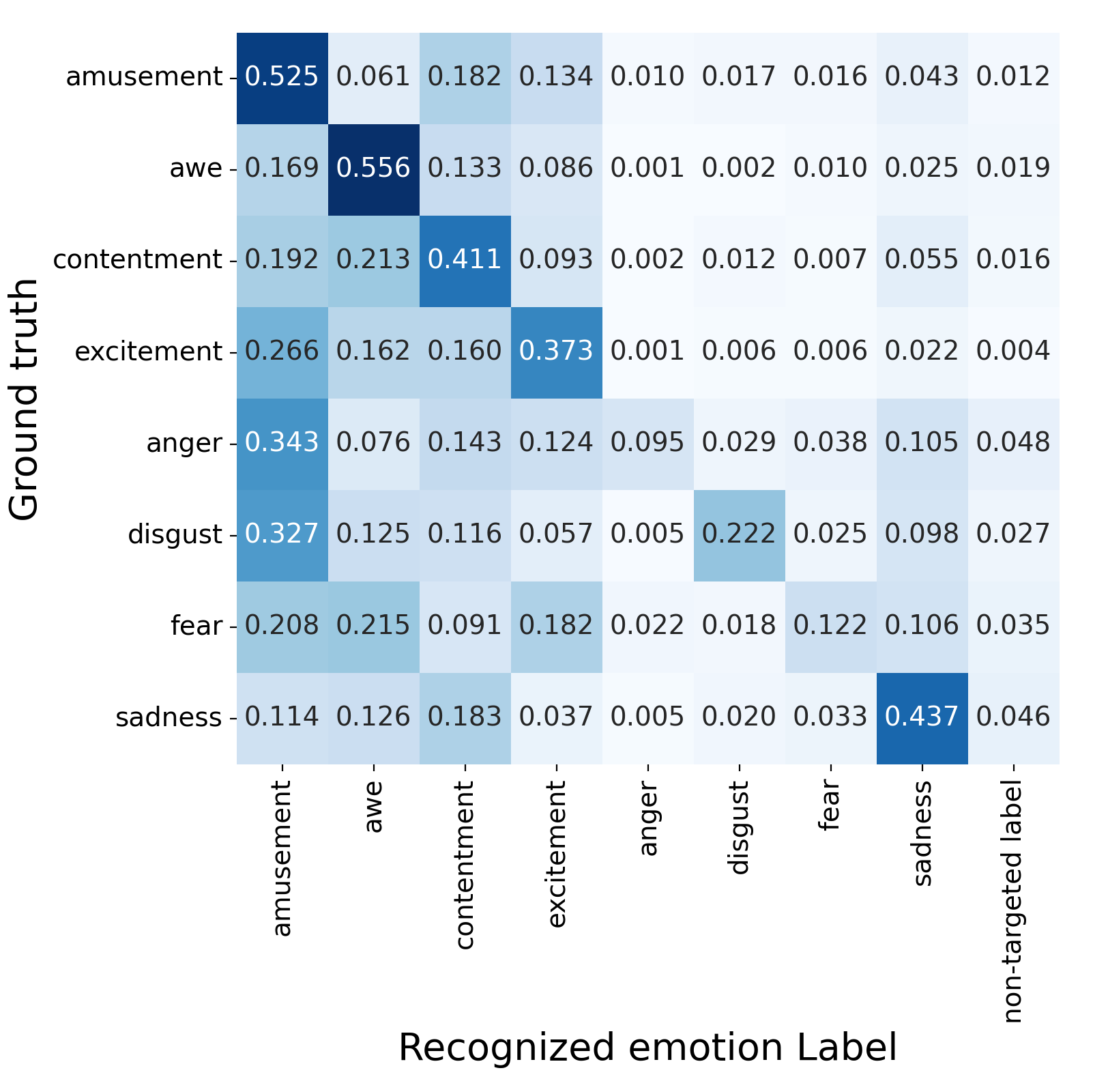}
    \vspace{-3mm}
    \caption{(ii) CM4}
    \label{fig:matrix_b}
  \end{subfigure}
  \caption{Confusion matrices of VER results obtained from  proposed method and CM4.}
  \label{fig:matrix}
\end{figure*}

\subsection{Qualitative experimental results}
TABLE~\ref{table2} shows the example prompts with the highest recognition accuracy in the training samples for different numbers of iterations. 
prompt~(i) is a simple command that only performs a VER task. 
In contrast, prompt (ii) indicates a specific role, and prompt (iii) follows a valid prompt representation for VER. 
This indicates that in the modification process, the LLM analyzes the valid parts of the prompts that are highly accurate and generates prompts suitable for the user. 
Finally, the use of prompt (iv) in the proposed method results in a more unique role for the model. 
From the recognition accuracy obtained using prompt (iv), we can confirm that this unique text representation enables the MLLM to extract emotional knowledge tailored to the user. 

Figure~\ref{fig:example1} shows examples of images, ground truth emotion labels, and recognition results obtained by the proposed method and ablation study cases: CMs 4--7. 
In image~1, compared with CM4 using the initial prompt, the proposed and other comparative methods recognized detailed information, such as facial expressions in person.
A reason for this is that methods other than CM4 modify the prompts to recognize more advanced information for VER.
In image~2, only the proposed method classified the image correctly.
Specifically, compared with CM5, which included a majority voting approach and discrete prompt tuning, a more accurate emotion recognition was achieved using the proposed method.
Images 3 and 4 show different emotion labels classified by different users.
Many of the comparative methods are not well personalized to the user and uniformly classify images. CM4 outputs an emotion other than the target emotion for user 5 because the prompt fails to convey the task correctly.
In contrast, the proposed method selects emotion labels tailored to each user.
These results confirm that the prompt used in the proposed method enables advanced consideration of the same object based on various elements of the image, not only the object.
Therefore, these examples demonstrate that discrete prompt tuning can adapt MLLM to personalized VER.

\subsection{limitations}
The proposed method still has several limitations. Figure~\ref{fig:matrix} shows the confusion matrices for the proposed method and CM4. 
Although the accuracy improves for most emotion labels, the performance for anger does not exhibit a noticeable improvement. 
As observed from the predicted emotion labels where the ground truth is anger, the results are extensively scattered across different categories. 
One possible reason is the class imbalance problem, because the number of anger images in the dataset is relatively small, resulting in biased learning and insufficient generalization. 
In future studies, it is necessary to design a model that can better leverage the knowledge of MLLMs to address class imbalance and achieve robust performance even with limited training samples.

\section{Conclusion}
\label{sec:conclusions}
This study proposes a personalized VER method based on
the discrete prompt tuning of an MLLM.
The proposed method enables the generation of user-specific discrete prompts that can be seamlessly applied to various black-box MLLMs. 
The experimental results demonstrate that these personalized prompts significantly improve VER performance. 
Notably, because the proposed method does not require access to model parameters or gradient information, it allows for the flexible integration of state-of-the-art, high-performance MLLMs without the need for additional fine-tuning. 
This approach not only enhances the adaptability and scalability of personalized VER systems but also opens up new possibilities for leveraging rapidly advancing MLLM technologies in diverse real-world applications.

\bibliographystyle{IEEEtran}
\bibliography{ref}

\begin{thebibliography}{10}
\providecommand{\url}[1]{#1}
\csname url@samestyle\endcsname
\providecommand{\newblock}{\relax}
\providecommand{\bibinfo}[2]{#2}
\providecommand{\BIBentrySTDinterwordspacing}{\spaceskip=0pt\relax}
\providecommand{\BIBentryALTinterwordstretchfactor}{4}
\providecommand{\BIBentryALTinterwordspacing}{\spaceskip=\fontdimen2\font plus
\BIBentryALTinterwordstretchfactor\fontdimen3\font minus \fontdimen4\font\relax}
\providecommand{\BIBforeignlanguage}[2]{{%
\expandafter\ifx\csname l@#1\endcsname\relax
\typeout{** WARNING: IEEEtran.bst: No hyphenation pattern has been}%
\typeout{** loaded for the language `#1'. Using the pattern for}%
\typeout{** the default language instead.}%
\else
\language=\csname l@#1\endcsname
\fi
#2}}
\providecommand{\BIBdecl}{\relax}
\BIBdecl

\bibitem{cnnrnn2017}
X.~Zhu, L.~Li, W.~Zhang, T.~Rao, M.~Xu, Q.~Huang, and D.~Xu, ``Dependency exploitation: a unified cnn-rnn approach for visual emotion recognition,'' in \emph{Proceedings of the International Joint Conference on Artificial Intelligence}, 2017, pp. 3595--3601.

\bibitem{xu2022mdan}
L.~Xu, Z.~Wang, B.~Wu, and S.~Lui, ``{MDAN}: Multi-level dependent attention network for visual emotion analysis,'' in \emph{Proceedings of the IEEE/CVF Conference on Computer Vision and Pattern Recognition}, 2022, pp. 9479--9488.

\bibitem{xie2024emovit}
H.~Xie, C.-J. Peng, Y.-W. Tseng, H.-J. Chen, C.-F. Hsu, H.-H. Shuai, and W.-H. Cheng, ``Emovit: Revolutionizing emotion insights with visual instruction tuning,'' in \emph{Proceedings of the IEEE/CVF Conference on Computer Vision and Pattern Recognition}, 2024, pp. 26\,596--26\,605.

\bibitem{zhu2025learning}
J.~Zhu, S.~Zhao, J.~Jiang, Z.~Xu, W.~Tang, and H.~Yao, ``Learning class prototypes for visual emotion recognition,'' in \emph{Proceedings of the IEEE International Conference on Acoustics, Speech and Signal Processing}, 2025, pp. 1--5.

\bibitem{poels2006capture}
K.~Poels and S.~Dewitte, ``How to capture the heart? reviewing 20 years of emotion measurement in advertising,'' \emph{Journal of Advertising Research}, vol.~46, no.~1, pp. 18--37, 2006.

\bibitem{lopez2017mining}
C.~E. Lopez and C.~S. Tucker, ``From mining affective states to mining facial keypoint data: The quest towards personalized feedback,'' in \emph{International Design Engineering Technical Conferences and Computers and Information in Engineering Conference}, vol. 58110, 2017, p. V001T02A039.

\bibitem{guntuku2019twitter}
S.~C. Guntuku, D.~Preotiuc-Pietro, J.~C. Eichstaedt, and L.~H. Ungar, ``What twitter profile and posted images reveal about depression and anxiety,'' in \emph{Proceedings of the international AAAI conference on web and social media}, vol.~13, 2019, pp. 236--246.

\bibitem{pang2015deep}
L.~Pang, S.~Zhu, and C.-W. Ngo, ``Deep multimodal learning for affective analysis and retrieval,'' \emph{IEEE Transactions on Multimedia}, vol.~17, no.~11, pp. 2008--2020, 2015.

\bibitem{guntuku2016likes}
S.~C. Guntuku, J.~T. Zhou, S.~Roy, W.~Lin, and I.~W. Tsang, ``Who likes what and, why?’insights into modeling users’ personality based on image ‘likes,'' \emph{IEEE Transactions on Affective Computing}, vol.~9, no.~1, pp. 130--143, 2016.

\bibitem{jaiswal2019intelligent}
S.~Jaiswal, S.~Virmani, V.~Sethi, K.~De, and P.~P. Roy, ``An intelligent recommendation system using gaze and emotion detection,'' \emph{Multimedia Tools and Applications}, vol.~78, pp. 14\,231--14\,250, 2019.

\bibitem{moroto2023zero}
Y.~Moroto, Y.~Ye, K.~Maeda, T.~Ogawa, and M.~Haseyama, ``Zero-shot visual sentiment prediction via cross-domain knowledge distillation,'' \emph{IEEE Open Journal of Signal Processing}, vol.~5, pp. 177--185, 2023.

\bibitem{kim2018building}
H.-R. Kim, Y.-S. Kim, S.~J. Kim, and I.-K. Lee, ``{Building Emotional Machines}: Recognizing image emotions through deep neural networks,'' \emph{IEEE Transactions on Multimedia}, vol.~20, no.~11, pp. 2980--2992, 2018.

\bibitem{li2019hierarchical}
L.~Li, X.~Zhu, Y.~Hao, S.~Wang, X.~Gao, and Q.~Huang, ``A hierarchical {CNN-RNN} approach for visual emotion classification,'' \emph{ACM Transactions on Multimedia Computing, Communications, and Applications}, vol.~15, no.~3s, pp. 1--17, 2019.

\bibitem{wu2023can}
C.~Wu, J.~Lei, Q.~Zheng, W.~Zhao, W.~Lin, X.~Zhang, X.~Zhou, Z.~Zhao, Y.~Zhang, Y.~Wang \emph{et~al.}, ``Can {GPT}-4v (ision) serve medical applications? case studies on {GPT}-4v for multimodal medical diagnosis,'' \emph{arXiv preprint arXiv:2310.09909}, 2023.

\bibitem{zhang2024mm}
C.~Zhang, K.~Lin, Z.~Yang, J.~Wang, L.~Li, C.-C. Lin, Z.~Liu, and L.~Wang, ``{MM-Narrator}: Narrating long-form videos with multimodal in-context learning,'' in \emph{Proceedings of the IEEE/CVF Conference on Computer Vision and Pattern Recognition}, 2024, pp. 13\,647--13\,657.

\bibitem{tzelepi2024disturbing}
M.~Tzelepi and V.~Mezaris, ``Disturbing image detection using {LMM}-elicited emotion embeddings,'' in \emph{Proceedings of the IEEE International Conference on Image Processing Challenges and Workshops}, 2024, pp. 4191--4196.

\bibitem{nadeem2024vision}
M.~Nadeem, S.~S. Sohail, L.~Javed, F.~Anwer, A.~K.~J. Saudagar, and K.~Muhammad, ``Vision-enabled large language and deep learning models for image-based emotion recognition,'' \emph{Cognitive Computation}, vol.~13, pp. 1--14, 2024.

\bibitem{lian2024gpt}
Z.~Lian, L.~Sun, H.~Sun, K.~Chen, Z.~Wen, H.~Gu, B.~Liu, and J.~Tao, ``{GPT}-4v with {Emotion}: A zero-shot benchmark for generalized emotion recognition,'' \emph{Information Fusion}, vol. 108, pp. 102\,367--102\,380, 2024.

\bibitem{bai2024m3d}
F.~Bai, Y.~Du, T.~Huang, M.~Q.-H. Meng, and B.~Zhao, ``{M3D}: Advancing {3D} medical image analysis with multi-modal large language models,'' \emph{arXiv preprint arXiv:2404.00578}, 2024.

\bibitem{zhang2024mathverse}
R.~Zhang, D.~Jiang, Y.~Zhang, H.~Lin, Z.~Guo, P.~Qiu, A.~Zhou, P.~Lu, K.-W. Chang, Y.~Qiao, P.~Gao, and H.~Li, ``{MATHVERSE}: Does your multi-modal {LLM} truly see the diagrams in visual math problems?'' in \emph{Proceedings of the European Conference on Computer Vision}, 2025, pp. 169--186.

\bibitem{lu2024gpt}
H.~Lu, X.~Niu, J.~Wang, Y.~Wang, Q.~Hu, J.~Tang, Y.~Zhang, K.~Yuan, B.~Huang, Z.~Yu, D.~He, S.~Deng, H.~Chen, Y.~Chen, and S.~Shan, ``{GPT} as psychologist? preliminary evaluations for {GPT}-4v on visual affective computing,'' in \emph{Proceedings of the IEEE/CVF Conference on Computer Vision and Pattern Recognition Workshops}, 2024, pp. 322--331.

\bibitem{mirchandani2023largelanguagemodelsgeneral}
S.~Mirchandani, F.~Xia, P.~Florence, B.~Ichter, D.~Driess, M.~G. Arenas, K.~Rao, D.~Sadigh, and A.~Zeng, ``Large language models as general pattern machines,'' in \emph{Proceedings of The Conference on Robot Learning}, vol. 229, 2023, pp. 2498--2518.

\bibitem{li2024formalityfavoredunravelinglearning}
J.~Li, Y.~Cao, S.~Huang, and J.~Chen, ``Formality is favored: Unraveling the learning preferences of large language models on data with conflicting knowledge,'' in \emph{Proceedings of the Conference on Empirical Methods in Natural Language Processing}, 2024, pp. 5307--5320.

\bibitem{sheng-etal-2019-woman}
E.~Sheng, K.-W. Chang, P.~Natarajan, and N.~Peng, ``The woman worked as a babysitter: On biases in language generation,'' in \emph{Proceedings of the Conference on Empirical Methods in Natural Language Processing and the International Joint Conference on Natural Language Processing}, 2019, pp. 3407--3412.

\bibitem{liu-etal-2022-p}
X.~Liu, K.~Ji, Y.~Fu, W.~Tam, Z.~Du, Z.~Yang, and J.~Tang, ``{P}-tuning: Prompt tuning can be comparable to fine-tuning across scales and tasks,'' in \emph{Proceedings of the Annual Meeting of the Association for Computational Linguistics}, 2022, pp. 61--68.

\bibitem{li2021prefix}
X.~L. Li and P.~Liang, ``Prefix-tuning: Optimizing continuous prompts for generation,'' in \emph{Proceedings of the Annual Meeting of the Association for Computational Linguistics and the International Joint Conference on Natural Language Processing}, 2021, pp. 4582--4597.

\bibitem{gu2021ppt}
Y.~Gu, X.~Han, Z.~Liu, and M.~Huang, ``{PPT}: Pre-trained prompt tuning for few-shot learning,'' in \emph{Proceedings of the Annual Meeting of the Association for Computational Linguistics}, 2022, pp. 8410--8423.

\bibitem{lester2021power}
B.~Lester, R.~Al-Rfou, and N.~Constant, ``The power of scale for parameter-efficient prompt tuning,'' in \emph{Proceedings of the Conference on Empirical Methods in Natural Language Processing}, 2021, pp. 3045--3059.

\bibitem{liu2024language}
S.~Liu, S.~Yu, Z.~Lin, D.~Pathak, and D.~Ramanan, ``Language models as black-box optimizers for vision-language models,'' in \emph{Proceedings of the IEEE/CVF Conference on Computer Vision and Pattern Recognition}, 2024, pp. 12\,687--12\,697.

\bibitem{diao2022black}
S.~Diao, Z.~Huang, R.~Xu, X.~Li, Y.~Lin, X.~Zhou, and T.~Zhang, ``Black-box prompt learning for pre-trained language models,'' \emph{arXiv preprint arXiv:2201.08531}, 2022.

\bibitem{pryzant-etal-2023-automatic}
R.~Pryzant, D.~Iter, J.~Li, Y.~Lee, C.~Zhu, and M.~Zeng, ``Automatic prompt optimization with ``gradient descent'' and beam search,'' in \emph{Proceedings of the Conference on Empirical Methods in Natural Language Processing}, 2023, pp. 7957--7968.

\bibitem{peng2015mixed}
K.-C. Peng, T.~Chen, A.~Sadovnik, and A.~C. Gallagher, ``A mixed bag of emotions: Model, predict, and transfer emotion distributions,'' in \emph{Proceedings of the IEEE/CVF Conference on Computer Vision and Pattern Recognition}, 2015, pp. 860--868.

\bibitem{you2016building}
Q.~You, J.~Luo, H.~Jin, and J.~Yang, ``Building a large scale dataset for image emotion recognition: The fine print and the benchmark,'' in \emph{Proceedings of the AAAI Conference on Artificial Intelligence}, vol.~30, no.~1, 2016.

\bibitem{chen2014deepsentibank}
T.~Chen, D.~Borth, T.~Darrell, and S.-F. Chang, ``{DeepSentiBank}: Visual sentiment concept classification with deep convolutional neural networks,'' \emph{arXiv preprint arXiv:1410.8586}, 2014.

\bibitem{you2015robust}
Q.~You, J.~Luo, H.~Jin, and J.~Yang, ``Robust image sentiment analysis using progressively trained and domain transferred deep networks,'' in \emph{Proceedings of the AAAI conference on Artificial Intelligence}, vol.~29, no.~1, 2015.

\bibitem{8825564}
D.~She, J.~Yang, M.-M. Cheng, Y.-K. Lai, P.~L. Rosin, and L.~Wang, ``{WSCNet}: Weakly supervised coupled networks for visual sentiment classification and detection,'' \emph{IEEE Transactions on Multimedia}, vol.~22, no.~5, pp. 1358--1371, 2020.

\bibitem{9964263}
S.~Deng, L.~Wu, G.~Shi, L.~Xing, W.~Hu, H.~Zhang, and Y.~Xiang, ``Simple but powerful, a language-supervised method for image emotion classification,'' \emph{IEEE Transactions on Affective Computing}, vol.~14, no.~4, 2023.

\bibitem{10388075}
C.~Bustos, C.~Civit, B.~Du, A.~Solé-Ribalta, and A.~Lapedriza, ``On the use of vision-language models for visual sentiment analysis: a study on clip,'' in \emph{Proceedings of the International Conference on Affective Computing and Intelligent Interaction}, 2023, pp. 1--8.

\bibitem{10889829}
B.~Wang, G.~Tu, B.~Liang, Z.~Bai, M.~Yang, X.~Zeng, L.~Yao, and R.~Xu, ``Enhancing emotion reasoning for image multi-emotion prediction,'' in \emph{Proceedings of the IEEE International Conference on Acoustics, Speech and Signal Processing}, 2025, pp. 1--5.

\bibitem{XU2024102366}
Q.~Xu, Y.~Wei, S.~Yuan, J.~Wu, L.~Wang, and C.~Wu, ``Learning emotional prompt features with multiple views for visual emotion analysis,'' \emph{Information Fusion}, vol. 108, p. 102366, 2024.

\bibitem{DBLP:journals/corr/abs-2103-00020}
A.~Radford, J.~W. Kim, C.~Hallacy, A.~Ramesh, G.~Goh, S.~Agarwal, G.~Sastry, A.~Askell, P.~Mishkin, J.~Clark, G.~Krueger, and I.~Sutskever, ``Learning transferable visual models from natural language supervision,'' in \emph{Proceedings of the International Conference on Machine Learning}, vol. 139, 2021, pp. 8748--8763.

\bibitem{deng2024learning}
S.~Deng, L.~Wu, G.~Shi, L.~Xing, M.~Jian, Y.~Xiang, and R.~Dong, ``Learning to compose diversified prompts for image emotion classification,'' \emph{Computational Visual Media}, vol.~10, no.~6, pp. 1169--1183, 2024.

\bibitem{dai2023instructblipgeneralpurposevisionlanguagemodels}
W.~Dai, J.~Li, D.~LI, A.~Tiong, J.~Zhao, W.~Wang, B.~Li, P.~N. Fung, and S.~Hoi, ``{InstructBLIP}: Towards general-purpose vision-language models with instruction tuning,'' in \emph{Proceedings of the Advances in Neural Information Processing Systems}, vol.~36, 2023, pp. 49\,250--49\,267.

\bibitem{cheng2024blackboxpromptoptimizationaligning}
J.~Cheng, X.~Liu, K.~Zheng, P.~Ke, H.~Wang, Y.~Dong, J.~Tang, and M.~Huang, ``Black-box prompt optimization: Aligning large language models without model training,'' \emph{arXiv preprint arXiv:2311.04155}, 2024.

\bibitem{achlioptas2023affection}
P.~Achlioptas, M.~Ovsjanikov, L.~Guibas, and S.~Tulyakov, ``{Affection}: Learning affective explanations for real-world visual data,'' in \emph{Proceedings of the IEEE/CVF Conference on Computer Vision and Pattern Recognition}, 2023, pp. 6641--6651.

\bibitem{mikels2005emotional}
J.~A. Mikels, B.~L. Fredrickson, G.~R. Larkin, C.~M. Lindberg, S.~J. Maglio, and P.~A. Reuter-Lorenz, ``Emotional category data on images from the international affective picture system,'' \emph{Behavior research methods}, vol.~37, no.~4, pp. 626--630, 2005.

\bibitem{korovina2019reliability}
O.~Korovina, M.~Baez, and F.~Casati, ``Reliability of crowdsourcing as a method for collecting emotions labels on pictures,'' \emph{BMC research notes}, vol.~12, pp. 1--6, 2019.

\bibitem{huang2024survey}
X.~Huang, J.~Xu, W.~Zheng, Q.~Mao, and A.~Dhall, ``A survey of deep learning for group-level emotion recognition,'' \emph{arXiv preprint arXiv:2408.15276}, 2024.

\bibitem{liu2024visual}
H.~Liu, C.~Li, Q.~Wu, and Y.~J. Lee, ``Visual instruction tuning,'' in \emph{Proceedings of the Advances in Neural Information Processing Systems}, vol.~36, 2023, pp. 34\,892--34\,916.

\bibitem{liu2021swintransformerhierarchicalvision}
Z.~Liu, Y.~Lin, Y.~Cao, H.~Hu, Y.~Wei, Z.~Zhang, S.~Lin, and B.~Guo, ``Swin transformer: Hierarchical vision transformer using shifted windows,'' in \emph{Proceedings of the IEEE/CVF International Conference on Computer Vision}, 2021, pp. 10\,012--10\,022.

\bibitem{zhao2024to}
C.~Zhao, J.~Shi, L.~Nie, and J.~Yang, ``To err like human: Affective bias-inspired measures for visual emotion recognition evaluation,'' in \emph{Proceeding of the Advances in Neural Information Processing Systems}, 2024, pp. 134\,747--134\,769.

\end{thebibliography}

\begin{IEEEbiography}[{\includegraphics[width=1in,height=1.25in,clip,keepaspectratio]{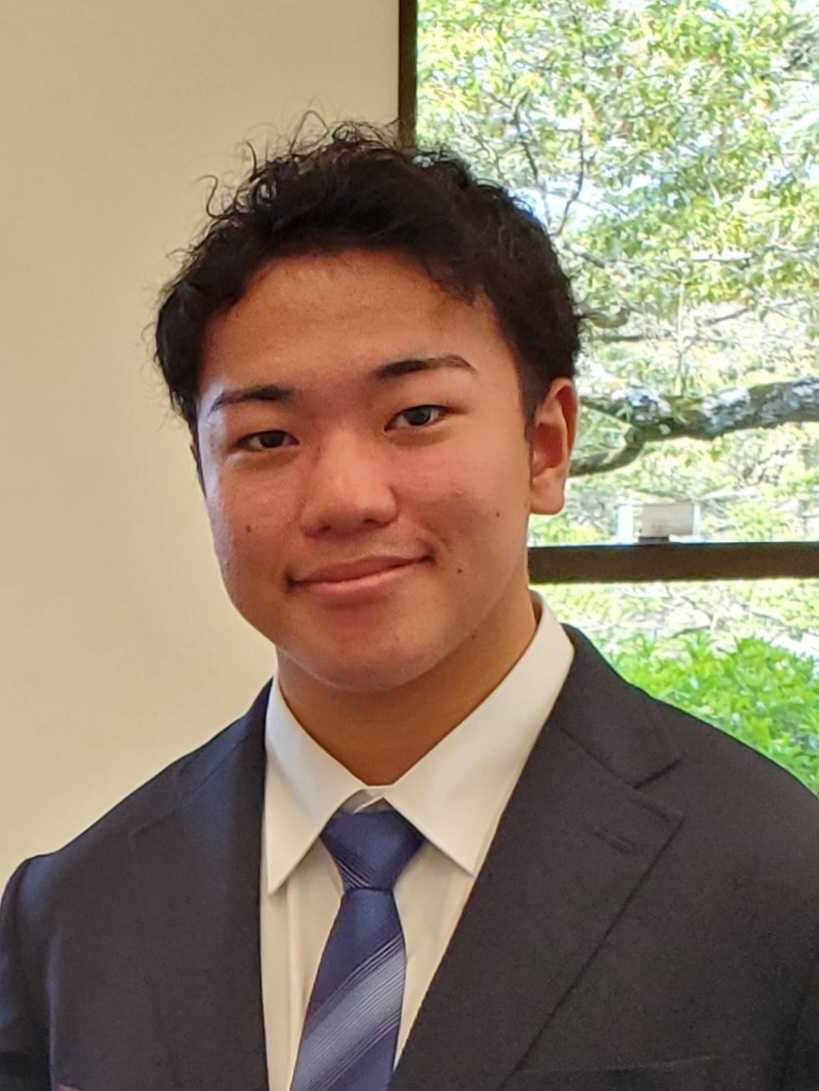}}]{Ryo Takahashi} (Graduate Student Member, IEEE) 
received his B.S. degree in electronics and information engineering from Doshisha University, Japan, in 2024, where he is currently pursuing his M.S. degree with the Graduate School of Information Science and Technology, Hokkaido University. 
His research interests include multimodal large language models and their applications.
He is a student member of IEICE and IEEE. 
\end{IEEEbiography}

\begin{IEEEbiography}[{\includegraphics[width=1in,height=1.25in,clip,keepaspectratio]{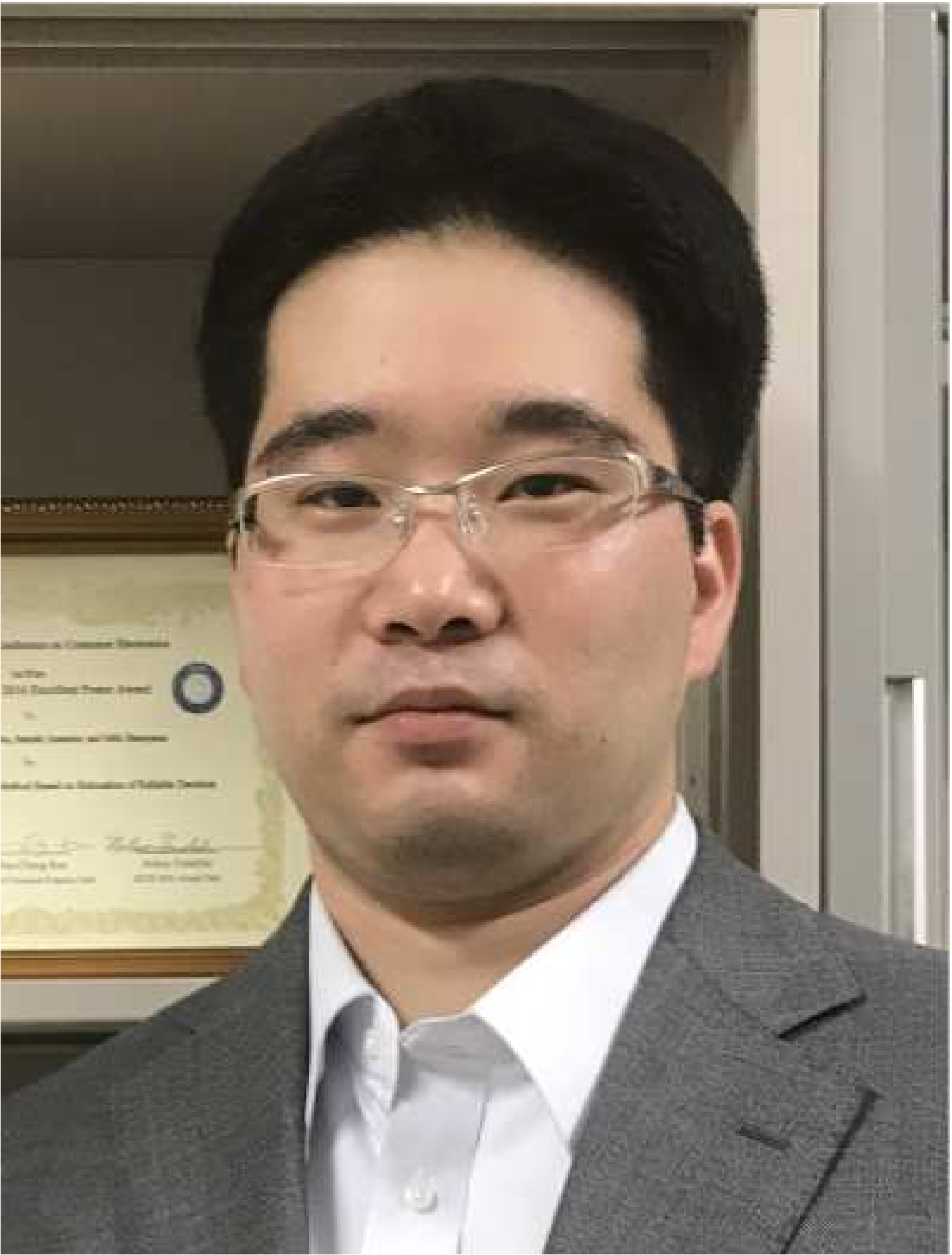}}]{Naoki Saito} (Member, IEEE) received his B.S. degree in electrical and electronic engineering from the National Institution for Academic Degrees and University Evaluation, Japan, in 2014, and his M.S. and Ph.D. degrees from Hokkaido University, Japan, in 2016 and 2019, respectively. He is currently an Assistant Professor with the Office of Institutional Research, Hokkaido University. His research interests include multimodal signal processing, machine learning, and their applications. He is a member of IEICE , JSAI, and IEEE. 
\end{IEEEbiography}

\begin{IEEEbiography}[{\includegraphics[width=1in,height=1.25in,clip,keepaspectratio]{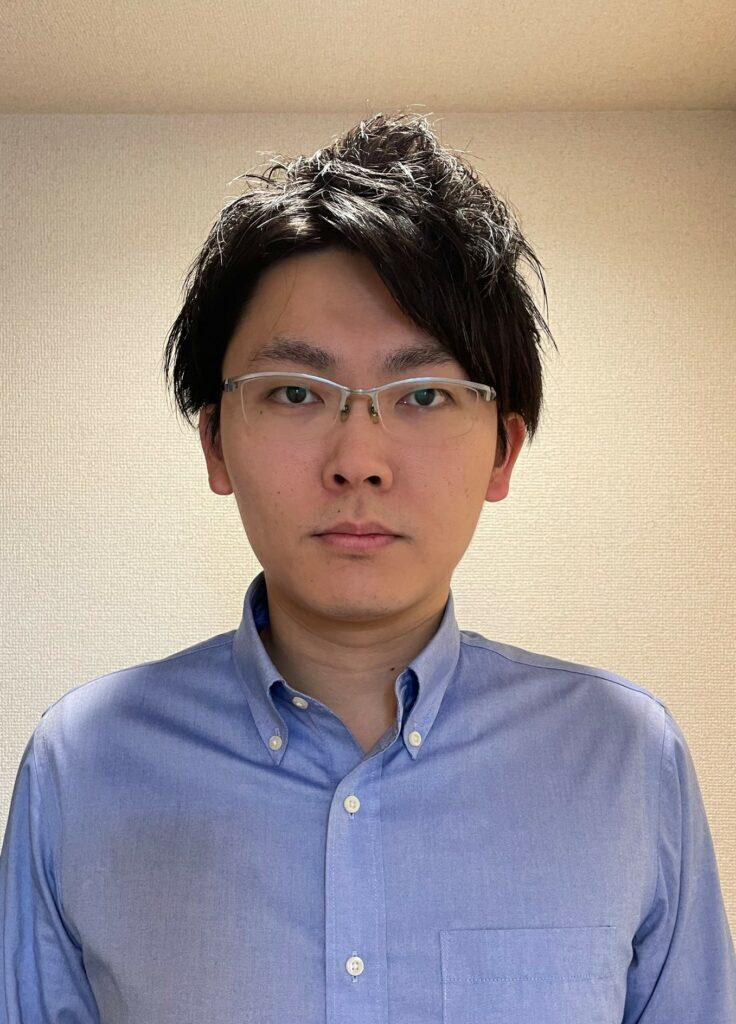}}]{Keisuke Maeda}
(Member, IEEE) 
received his B.S., M.S., and Ph.D. degrees in electronics and information engineering from Hokkaido
University, Japan, in 2015, 2017, and 2019, respectively. 
He is currently an Associate Professor with the Faculty of Information Science and Technology, Hokkaido University. 
His research interests include multimodal signal processing and machine learning and its applications. 
He is a member of IEICE and IEEE.
\end{IEEEbiography}

\begin{IEEEbiography}[{\includegraphics[width=1in,height=1.25in,clip,keepaspectratio]{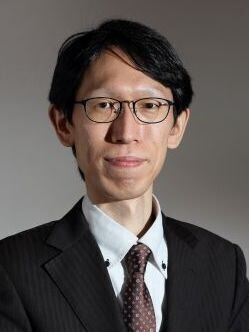}}]{Takahiro Ogawa} (Senior Member, IEEE)  received his B.S., M.S., and Ph.D. degrees in electronics and information engineering from Hokkaido University, Japan, in 2003, 2005, and 2007, respectively. He joined the Graduate School of Information Science and Technology, Hokkaido University, in 2008. He is currently a Professor with the Faculty of Information Science and Technology, Hokkaido University. His research interests include artificial intelligence, the Internet of Things, big data analysis for multimedia signal processing, and its applications. He is a member of ACM, IEICE, ITE, and IEEE. He was the Special Session Chair of IEEE ISCE2009, the Doctoral Symposium Chair of ACM ICMR2018, the Organized Session Chair of IEEE GCCE2017-2019, the TPC Vice Chair of IEEE GCCE2018, and the Conference Chair of IEEE GCCE2019. Furthermore, he has been an Associate Editor of \textit{ITE Transactions on Media Technology and Applications}.
\end{IEEEbiography}

\begin{IEEEbiography}[{\includegraphics[width=1in,height=1.25in,clip,keepaspectratio]{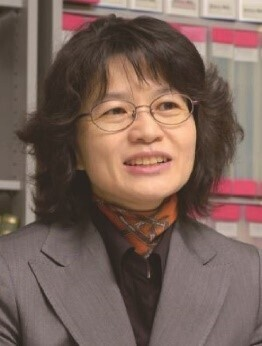}}]{Miki Haseyama} (Senior Member, IEEE)  received her B.S., M.S., and Ph.D. degrees in electronics from Hokkaido University, Japan, in 1986, 1988, and 1993, respectively. She joined the Graduate School of Information Science and Technology, Hokkaido University, as an Associate Professor, in 1994. She was a Visiting Associate Professor with Washington University, USA, from 1995 to 1996. She is currently a Professor with the Faculty of Information Science and Technology, Hokkaido University. Her research interests include image and video processing and its development into semantic analysis. She is a fellow of ITE and a member of IEICE, ASJ, and IEEE. She has been the Vice-President of the Institute of Image Information and Television Engineers (ITE), Japan, an Editor-in-Chief of ITE Transactions on Media Technology and Applications, and the Director of International Coordination and Publicity at the Institute of Electronics, Information and Communication Engineers (IEICE).

\end{IEEEbiography}

\EOD

\end{document}